\documentclass[12pt]{article}
\usepackage[T1]{fontenc}
\usepackage[latin9]{inputenc}
\usepackage{geometry}
\usepackage{float}
\usepackage{amsmath}
\usepackage{amssymb}
\usepackage{graphicx}
\usepackage{bbm}
\usepackage{times}
\allowdisplaybreaks

\begin{document}

\title{Calibrating Black Box Classification Models through the Thresholding Method}
\author{Arun Srinivasan}
\date{May 20, 2017}
\maketitle

\begin{abstract}
\footnotesize{\noindent In high-dimensional classification settings, we wish to seek a balance between high power and ensuring control over a desired loss function. In many settings, the points most likely to be misclassified are those who lie near the decision boundary of the given classification method. Often, these uninformative points should not be classified as they are noisy and do not exhibit strong signals. In this paper, we introduce the Thresholding Method to parameterize the problem of determining which points exhibit strong signals and should be classified. We demonstrate the empirical performance of this novel calibration method in providing loss function control at a desired level, as well as explore how the method assuages the effect of overfitting. We explore the benefits of error control through the Thresholding Method in difficult, high-dimensional, simulated settings. Finally, we show the flexibility of the Thresholding Method through applying the method in a variety of real data settings.}
\end{abstract}

\section{Introduction}
\noindent In classification problems, the most difficult points to classify are those that lie near the classification margins. In high-dimensional settings, this problem is further exacerbated with a majority of the misclassification error being attributed to those points. Further, in high-dimensional settings we often do not know if our chosen fitting method accurately models the unknown, underlying data generating distribution. In these cases, it is difficult to calibrate the model to account for this mismatch. While a well-fit model can assuage these issues to a point, problems such as overfitting can make it difficult to ascertain the quality of the fit on the testing data.

\vspace{4mm} \noindent Often, these points on the margins are not informative as it is difficult to determine which label they originate from. For example, in the case of Gaussian Mixture Models, which we will explore further in this paper, the hardest to classify points are those who do not appear to belong strongly to any mixture component. Therefore, it is often desired to determine which points are the most likely to be correctly classified, and what proportion of the underlying dataset should be classified to achieve a desired error rate. While the method we propose is agnostic to the underlying model used to fit the data, we will use the Gaussian Mixture Model in the simulated and real data analysis.

\vspace{4mm} \noindent Low training error alone does not provide suitable control over the error expected in the testing set. To resolve this, we propose the Thresholding Method to control the conservativeness of the fitted classifier in classifying these points. By using a hold-out set to tune an introduced parameter, we demonstrate that this method provides a mechanism of calibration over a target error function on the testing set.

\vspace{4mm} \noindent We first provide a brief background on the Gaussian Mixture Model classifier, as well as demonstrate how the classifier fulfills the requirements for the Thresholding Method. By discussing the Maximum Likelihood Classifier, we detail how the likelihood function can be used as a score function for calibration.
\section{Background}
\vspace{4mm} \noindent The Thresholding Method is a flexible calibration algorithm that can be applied to a wide variety of classification methods. We apply the Thresholding Method after fitting the data with any black box classifier that outputs a set of class labels for each point as well as a score for each point that represents our confidence in classifying the point to each possible label.

\vspace{4mm} \noindent As mentioned previously, we apply the Thresholding Method to the commonly used Gaussian Mixture Model and Maximum Likelihood Classifier to demonstrate the effectiveness of the method. However, any classifier that outputs both the required labels and associated scores may be used. For example, a SVM could be used to determine class labels as well as providing a score for the SVM classification. This would be satisfactory for use with the Thresholding Method.

\vspace{4mm} \noindent As stated previously, the Gaussian Mixture Model framework and the Maximum Likelihood Classifier were chosen for exploration as they are straight-forward and easy to interpret.  However, as we will show through real data and simulation results, the Thresholding Method is a useful tool to calibrate even these simple frameworks, and can be extended to more complicated techniques.

\subsection{Gaussian Mixture Models}
\vspace{4mm} \noindent The Gaussian Mixture Model setting arises in many application areas, such as genetics and discourse analysis. In this setting, the data $\mathbf{X}\in\mathbb{R}^{n\times d}$ are generated from a mixture of $k$ Gaussian models. Each of the data points $\mathbf{X}_i \in \mathbb{R}^{1\times d}$ is generated from a specific Gaussian distribution, with a individual mean $\mu_j \in \mathbb{R}^{1\times d}$ and covariance matrix $\Sigma_j\in \mathbb{R}^{d\times d}$ where $j\in\{1,...,k\}$. The mixture proportions $\pi_j$ indicate the relative weights of each of the individual Gaussian distributions in the full mixture model. The form of the Gaussian Mixture Model is as follows:

$$f(\mathbf{X}_i)=\sum_{j=1}^{k}\pi_j\times \mathcal{N}(\mathbf{X}_i|\mu_j,\Sigma_j)$$

\noindent In this setting, we explore the supervised learning case, were we are given the latent labels $Z$ for each of the data points where $Z$ can take values in $\{1,...,k\}$. Let $n_j$ be the number of points assigned to label $j$. Given these labels,  the sample Gaussian distribution means, covariance matrices, and mixture proportion for components $j=\{1,...,k\}$ can easily be estimated as follows:

\vspace{4mm} \noindent \textit{Estimation of Component Means:} 
$$\mu_j=\frac{\sum_{i=1}^{n} \mathbf{X}_i \mathbbm{1}\{Z_i = j\}}{n_j}$$
\\
\\
\vspace{4mm}\noindent\textit{Estimation of Component Covariance Matrices:} 
\\
\vspace{4mm}For each $j\in\{1,...,k\}$, let $\mathbf{Y}_j$ be the rows of $\mathbf{X}$ where $Z_i = j$:
$$\Sigma_j = \frac{1}{n_j-1}\mathbf{Y}_j^\intercal \mathbf{Y}_j$$

\noindent \textit{Estimation of Component Mixtures:} 
$$\pi_j=\frac{\sum_{i=1}^{n} \mathbbm{1}\{Z_i = j\}}{n}$$

\vspace{4mm} \noindent Several methods can be used for the classification problem such as $\textit{Maximum Likelihood Classification}$ (Bayes Classifier) or methods such as $\textit{Support Vector Machines}$ (SVMs), and higher complexity methods.

\subsection{Maximum Likelihood Classification}
The Maximum Likelihood Classification method involves classifying a new point to the mixture component that produces the largest likelihood. The Maximum Likelihood Classifier can be used to classify points from a variety of fit models; however, for the purposes of exploration in this paper we mainly focus on classifying points generated from a Gaussian Mixture Model. However, we emphasize that the Maximum Likelihood Classifier can be used for many fitting models, by simply changing the likelihood function used. We also demonstrate that the Maximum Likelihood Classifier satisfies the requirements for applying the Thresholding Method.

\vspace{4mm} \noindent When applied to the Gaussian Mixture Model framework, as we have estimated the mixture means, covariance matrices, and mixture proportions, we are able calculate the likelihoods, $L^{new}_j$ for $j\in\{1,...,k\}$ of the new point $X_{new}$ for each of the mixture components $j\in\{1,...,k\}$ as follows:

$$L^{new}_j = P(X_{new}|Z_{new}=j) = \pi_j \times f(X_{new}|\mu_j, \Sigma_j)$$

\noindent Where, $f$ is distributed Gaussian with mean and covariance matrix $\mu_j$ and $\Sigma_j$ respectively.

\vspace{4mm} \noindent Regardless of the underlying model fit, given these likelihoods, we are then able to determine which label to classify the point to by choosing the label which maximizes the likelihood.

$$ Z_{new} = \underset{j}{\text{argmax}} \: L^{new}_j $$

\noindent For computational purposes, we tend to prefer working on the log-likelihood scale and maximizing $\ell^{new}_j=\log(L^{new}_j) $. So, 

$$ Z_{new} = \underset{j}{\text{argmax}} \: \ell^{new}_j $$

\vspace{4mm} \noindent The Gaussian Mixture Model along with the Maximum Likelihood Classifier satisfies the requirements for calibration through the Thresholding Method. After fitting the underlying model as a Gaussian Mixture, applying the Maximum Likelihood Classifier outputs a set of labels for which component each data point most likely originated from.

\vspace{4mm} \noindent Next, the Thresholding Method requires a set of scores enumerating our confidence in classifying a point to each possible label. In this classifier, the log-likelihood of each point for each possible label functions as a score. Higher log-likelihood values indicate that a point is more likely to have been generated from a given mixture component. As such we are able to rank the labels in order of log-likelihood.

\section{Thresholding Method for Classification}

\subsection{Misclassification Rate Formulation}

\noindent In many classification methods, the points that are misclassified lie on the boundary between multiple classification regions. For example, in the case of SVMs, the misclassified points tend to lie in the margin around the separating hyperplane. The further a point is from the boundaries, the more confident we are classifying an individual point. 

\vspace{4mm} \noindent The objective of the Thresholding Method is to determine which points should be classified to guarantee a Misclassification Rate (MR) $\leq q$, where $q$ is the target MR. The Misclassification Rate is the expected value of the Misclassification Proportion (MCP). For ease of notation we will use the two interchangeably in the simulation and real data analysis sections.

\vspace{4mm} \noindent \textit{Misclassification Rate (MR)}:
$$\text{MR} = \mathbb{E}[\text{MCP}]=\mathbb{E}[\frac{\sum_{i = 1}^{n} \mathbbm{1}\{Z_i\not=Z^\star_i\}\times \mathbbm{1}\{i\in C\}}{|C|}]$$

\vspace{4mm} \noindent \textit{Probability of Assignment (PA)}:
$$\text{PA}=\frac{|C|}{n}$$

\noindent Where $C\subseteq \{1,2,...,n\}$ is the set of points classified and $Z^\star$ are the true labels for each point.

\vspace{4mm} \noindent Through the Thresholding Method, we introduce the parameter $t$ which defines how much more certain we must be on assigning a point to the most likely group. We then sweep $t$ from a lax threshold, where we assign all points, to a strict threshold, where we must be much more stringent on what points we assign. At each value of $t$ we run our desired classification method and calculate both the MCP and PA on the validation dataset. PA represents the effect of changing $t$, as higher $t$ values monotonically decrease the number of points classified. Then, by inspecting the two metrics on the hold-out dataset, we are then able to determine which value of $t$ to choose such that the MCP at $t$ is $\leq q$. Let the $t$ that satisfies the previous condition be known at $t^*$. Given $t^*$, we then are able to determine approximately how many points will be classified by mapping to the associated PA. Further, we then use the $t^*$ as the thresholding level on the testing data to classify new points.

\vspace{4mm} \noindent In the Maximum Likelihood Classification scenario, the Thresholding Method is as follows:

 {\normalsize \vspace{4mm} \noindent \textbf{ \textsc{Thresholding Method for Target MCP}}}

\vspace{4mm} \noindent Let $\mathbf{X}^{tr}\in \mathbb{R}^{n_{tr}\times d}$ be the labeled training data with true labels $Z_{tr}^*$, $\mathbf{X}^{v}\in\mathbb{R}^{n_{v}\times d}$ be the validation data with labels $Z_{v}^*$ , and $\mathbf{X}^{te}\in \mathbb{R}^{n_{te}\times d}$ be the testing data. Let $T=\{t_{min},...,t_{max}\}$ be the set of thresholds to evaluate. Let $\mathbf{MCP}$ be the vector of MCPs at each $t\in T$ and let $\mathbf{PA}$ be the vector of PAs for each $t\in T$. Let $\mathbf{Z}\in\mathbb{N}_0^{n_v\times |T|}$ denote the matrix of labels for each point at each threshold level. For shorthand let $\mathbf{Z}_{i,m} = 0$ indicate that point $i$ is not classified at a given threshold level $m$. 
\begin{enumerate}
\item Using $\mathbf{X}^{tr}$ and $Z_{tr}^*$, fit the selected underlying model. For example, in the Gaussian Mixture Model scenario, fit the maximum likelihood estimates for mixture means, covariance matrices and mixture components.
\item For each $\mathbf{X}^{v}_i$ calculate the $s_{i,j}$, where $s_{i,j}$ is the score output by the classifier for point $\mathbf{X}^{v}_i$ for label $j$. For example, in the Maximum Likelihood Classifier, $s_{i,j} = \ell_{i,j}$.
\item For each $m \in \{1,...,|T|\}$
\begin{enumerate}
\item For each $i \in \{1,2,...,n\}$
\begin{enumerate}
\item Let $s_i^{max} = \underset{j}{\text{argmax}} \: s_{i,j}$ with the maximizing label as $j_{max}$ for the highest score.
\item Let $s_i^{second} = \underset{j\in\{1,..,k\}/\{j_{max}\}}{\text{argmax}} \: s_{i,j}$ denote the second highest score.
\item If $s_i^{max} - s_i^{second} \geq T_m$
\begin{enumerate}
\item $\mathbf{Z}_{i,m}=j_{max}$
\end{enumerate}
\item If $s_i^{max} - s_i^{second} < T_m$
\begin{enumerate}
\item $\mathbf{Z}_{i,m}= 0$
\end{enumerate}
\end{enumerate}
\item Calculate $\mathbf{MCP}_m$ and $\mathbf{PA}_m$ given $\mathbf{Z}_{\cdot,m}$ and true labels $Z^*$.
\end{enumerate}
\item To determine what thresholding level to use to ensure $MR \leq q$, calculate $t^* = T_{m^*}$ where $m^*$ is the $m$ corresponding to largest entry of $\mathbf{MCP}$ that is below $q$.
\item For each $\mathbf{X}^{te}_i$, use the classifier to calculate associated scores $s_{i,j}$ for the testing data and repeat 3(a) and 3(b) at thresholding level $t^*$ to classify testing data.
\end{enumerate}

\vspace{4mm} \noindent As seen in the method above, the strictness of the classification scheme is controlled by $t$ as desired. The larger $t$ is, the larger the gap between the highest and second highest score components must be. Therefore, as $t$ increases, the number of points that are classified decreases, as we only classify the points that we believe to show a strong signal towards a specific label. These are points that are further away from the decision boundaries. Not only does this method provide control of the MR, but also assists us in determining how difficult our classification problem is through examining the PA. If for a reasonable target $q$ the associated PA at that threshold level is low, this indicates that many points are near classification boundaries thereby leading to a difficult problem, which is common in a higher-dimensional setting.

\subsection{Soft-max Thresholding Formulation}
The Thresholding Method is flexible, capable of being applied to different metrics other than MR and PA. We extend the previous formulation to the case where we want to control the certainty in our classification scheme. In the Maximum Likelihood Classification Scheme, we can measure our certainty in our classification through applying the softmax function and weighting the log-likelihoods by the thresholds. Let $s=\frac{1}{t}$ where $t$ is a given threshold. Further, for a point $\mathbf{X}_i$, with true label $Z_i$ we can calculate the Soft-max Threshold Probability for each label $j \in\{1,...,k\}$ as:

\vspace{4mm} \noindent \textit{Soft-max Threshold Probabilities}:

$$p(\mathbf{X}_i)_j^s = \frac{e^{s \ell_{i,j}}}{\sum_{m=1}^{k} e^{s \ell_{i,m}}}$$

\vspace{4mm} \noindent By the formulation above, as $t$ increases, the probabilities are distributed evenly across each of the potential mixture components. This indicates how certain we are in our classification, as for the probability of a component to be large, it must be larger than the thresholding level. If the thresholding level is extremely large in comparison to the largest log-likelihood, the probability distribution for the point across all components becomes uniform.

\vspace{4mm} \noindent Instead of focusing on MR and PA, we can choose new metrics which focus on giving information on the uncertainty and information in our classification scheme. The Multinomial Classification Loss (MCL) and Average Entropy (AE) were chosen as metrics and are defined as follows:

\vspace{4mm} \noindent \textit{Multinomial Classification Loss (MCL)}:
$$MCL_i^s =-\sum_{i=1}^n\sum_{j=1}^K\mathbbm{1}\{Z_i =j\}\log(p(\mathbf{X}_i)_j^s)$$
\vspace{4mm} \noindent \textit{Average Entropy (AE)}:
$$AE_i^s = \frac{-\sum_{i=1}^{n} \sum_{j=1}^k \log(p(\mathbf{X}_i)_j^s)}{n} $$

\vspace{4mm} \noindent Therefore, if $t$ is extremely large, the probabilities are uniform across the $k$ labels. The maximum value that AE can take is $-\log(\frac{1}{k})$.

\vspace{4mm} \noindent In this scheme, instead of indicating a target MCP, we instead can indicate that we desire a target MCL of at most $r$. Using this $r$ we can run a similar Thresholding Method as in the MCP formulation detailed above.

{\normalsize {\vspace{4mm} \noindent \textbf{\textsc{Thresholding Method for Target MCL}}}}

\vspace{4mm} \noindent \noindent Let $\mathbf{X}^{tr}\in \mathbb{R}^{n_{tr}\times d}$ be the labeled training data with true labels $Z_{tr}^*$, $\mathbf{X}^{v}\in \mathbb{R}^{n_{v}\times d}$ be the validation data with labels $Z_{v}^*$ , and $\mathbf{X}^{te}\in \mathbb{R}^{n_{te}\times d}$ be the testing data. Let $T=\{t_{min},...,t_{max}\}$ be the set of thresholds to evaluate. Let $\mathbf{MCL}$ be the vector of MCLs at each $t\in T$ and let $\mathbf{AE}$ be the vector of AEs for each $t\in T$. Let $\mathbf{Z}\in\mathbb{N}_0^{n_v\times |T|}$ denote the matrix of labels for each point at each threshold level. For shorthand let $\mathbf{Z}_{i,m} = 0$ indicate that point $i$ is not classified at a given threshold level $m$.

\vspace{4mm} \noindent Let $\mathbf{P}\in [0,1]^{n\times k \times |T|}$ be the matrix of soft-max probabilities for each point, across all possible classification values, at each threshold level.

\begin{enumerate}
\item Using $\mathbf{X}^{tr}$ and $Z_{tr}^*$, fit the selected underlying model. For example, in the Gaussian Mixture Model scenario, fit the maximum likelihood estimates for mixture means, covariance matricies and mixture components.
\item For each $\mathbf{X}^{v}_i$ calculate the $s_{i,j}$, where $s_{i,j}$ is the score output by the classifier for point $\mathbf{X}^{v}_i$ for label $j$. For example, in the Maximum Likelihood Classifier, $s_{i,j} = \ell_{i,j}$.
\item For each $m \in \{1,...,|T|\}$
\begin{enumerate}
\item For each $i \in \{1,2,...,n\}$
\begin{enumerate}
\item Let $\ell_i^{max} = \underset{j}{\text{argmax}} \: \ell_{i,j}$ with the maximizing label as $j_{max}$ for the highest log-likelihood. 
\item Let $\ell_i^{second} = \underset{j\in\{1,..,k\}/\{j_{max}\}}{\text{argmax}} \: \ell_{i,j}$ denote the second highest log-likelihood.
\item If $\ell_i^{max} - \ell_i^{second} \geq T_m$
\begin{enumerate}
\item $\mathbf{Z}_{i,m}=j_{max}$
\end{enumerate}
\item If $\ell_i^{max} - \ell_i^{second} < T_m$
\begin{enumerate}
\item $\mathbf{Z}_{i,m}= 0$
\end{enumerate}
\item For each $j\in \{1,...,k\}$
\begin{enumerate}
\item $\mathbf{P}_{i,j,m}=p(\mathbf{X}_i)_j^{\frac{1}{m}}$
\end{enumerate}
\end{enumerate}
\item Calculate $\mathbf{MCL}_m$ given labels $Z_v^*$ and $\mathbf{P}_{\cdot, \cdot, m}$.
\item Calculate $\mathbf{AE}_m$ given $\mathbf{P}_{\cdot, \cdot, m}$
\end{enumerate}
\item To determine what thresholding level to use to ensure $MCL \leq r$, calculate $t^* = T_{m^*}$ where $m^*$ is the $m$ corresponding to largest entry of $\mathbf{MCL}$ that is below $r$.
\item For each $\mathbf{X}^{te}_i$, calculate the $\ell_{i,j}$ and repeat 3(a), 3(b) and 3(c) at thresholding level $t^*$ to classify testing data.
\end{enumerate}
\noindent Once again, this method involves using $t$ to parameterize the strictness of the problem. When $t$ is large, consequently $s$ is extremely small. Therefore, the probabilities are uniformly spread across each of the $p(\mathbf{X}_i)_j^s$.  As $t$ shrinks, $s$ increases, yielding a more lax classification scheme. The $p(\mathbf{X}_i)_j^s$ tend to become peaked around the components where the likelihood is the largest. In the case of a very small $t$, this tends to place all the probability mass on a single component. In this case, the AE is very low; however, the MCL becomes large. Therefore, by designating a desired MCL level $r$ and running the method on our hold-out data to determine $t^*$, we can determine the AE we expect to see in the testing set and ensure that the loss near what we desire on the testing set.

\subsection{Flexibility of the Thresholding Method}
 \noindent While the metrics of interest differ between MCP formulation and the MCL formulation, the mechanics of the Thresholding Method remain the same. The flexibility of the Thresholding method affords minimal restrictions on the metrics of interest.

\vspace{4mm} \noindent We simply require the metric we wish to control, MCP or MCL in these formulations, to be generally monotonic, though we allow them to not be smooth. In both of these cases, AE and PA represent the effect of the single parameter introduced by the Thresholding Method, $t$. The metric that represents the effect of $t$ only needs to be monotonic. While both AE and PA measure the direct effect of $t$, they both interpret $t$ in different ways. PA measures conservativeness by measuring the number of points of data classified, while AE operates on all points, not simply the classified points, to measure the certainty of the method. As $t$ increases, the AE monotonically increases, while the PA monotonically decreases. These monotonicity requirements are necessary for the selection of $t^*$, as we select the $t^*$ corresponding to the largest entry of our desired error function that is below the selected level.

\vspace{4mm} \noindent While the two formulations are presented as different algorithms, the framework of the Thresholding Method renders them to be variations on the underlying calibration scheme. As long as these two requirements hold, the Thresholding Method can be applied to a variety of different metrics. When coupled with the minimal requirements on the black box underlying classification method, we believe the Thresholding Method can be used in a wide variety of situations and in many application areas. 
\section{Simulation Results}
\subsection{Simulation Results}
We demonstrate the effectiveness of the thresholding method on simulated Gaussian Mixture data. In particular, we will analyze the method across different levels of problem difficulty. As the difficulty of the classification problem increases, more overfitting tends to take place. We demonstrate that as the problem difficulty increases, the Thresholding Method performs better than the naive method in ensuring desired misclassification rates.
\subsubsection{Difficulty from Separation of Mixture Components}
 \noindent We investigate two settings of classification problem difficulty. The first is difficulty caused through the overlap of the mixture components. An easy problem in this context is where the mixture components are well separated from one another. The difficulty of the problem increases, as the overlap between components increases. 

\vspace{4mm} \noindent In this scenario, we generate data from a mixture model with two components with even mixture proportions. The covariance matrices for each component were identical and diagonal. The diagonal entries of each covariance matrix were set to .5. In this scenario, we have selected a 30 dimensional problem to simulate a high dimensional setting. We generated a total of $n = dim\times 5$ points to be split into the training, testing, and hold-out datasets. The data was split as follows with $n_{train} = 90$, $n_{hold} = 30$, and $n_{test} = 30$. 

\vspace{4mm} \noindent The difficulty of the problem was controlled by the distance between the centers of the two mixture components. In the harder difficulty problems, the centers were brought near each other to make it much more difficult to label each class. To ensure that problem was difficult and to have increased control over location of the centers, the centers were randomly generated in two dimensions, while the value for the remaining coordinates were set to zero. Therefore, the centers lie on a plane within the higher dimensional space. 
\begin{figure}[H]
\begin{centering}
{\includegraphics[scale=0.35]{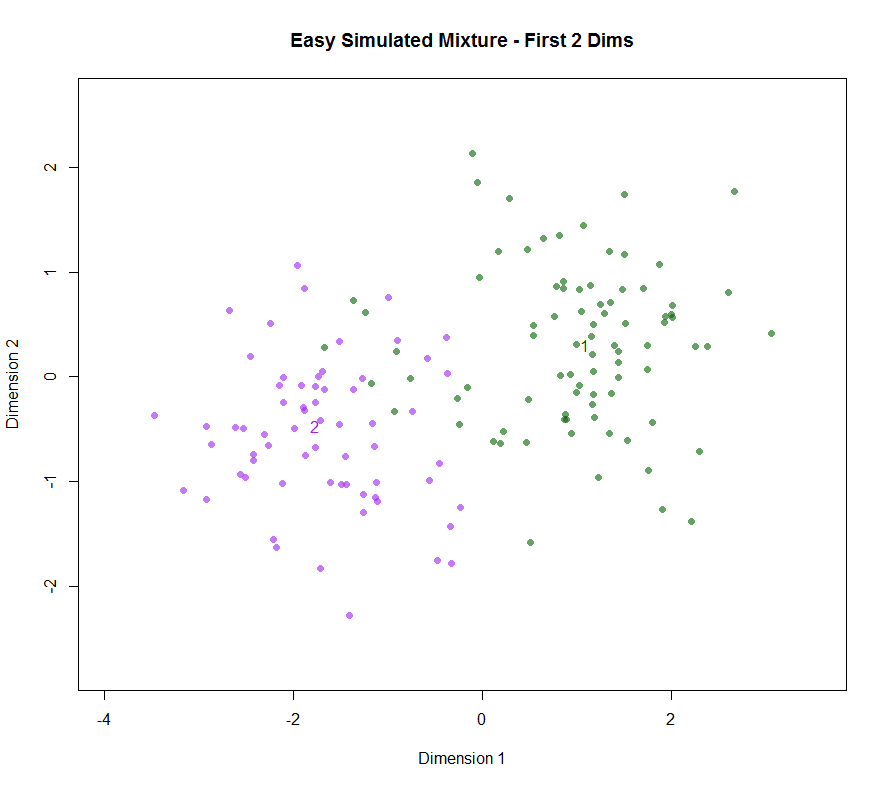}}{\includegraphics[scale=0.35]{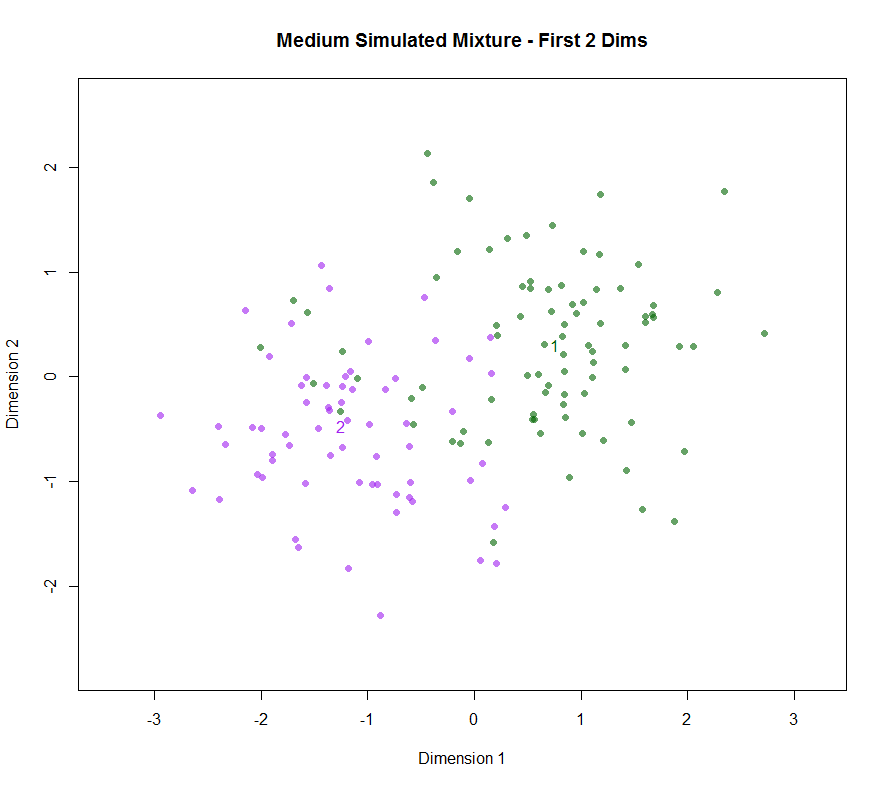}}
{\includegraphics[scale=0.35]{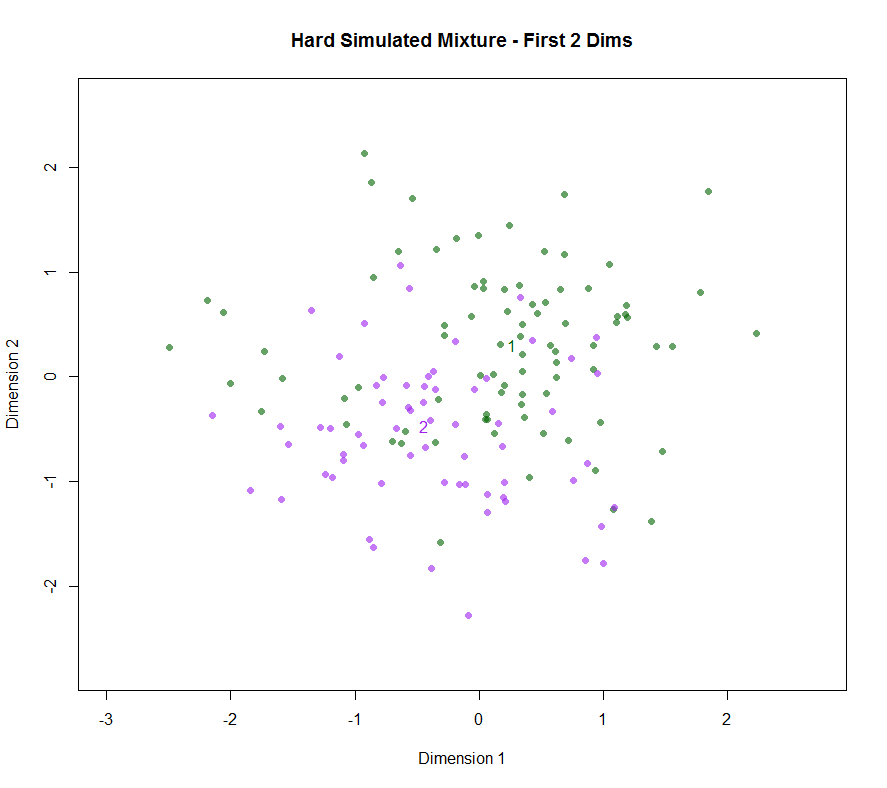}}
\par\end{centering}
\caption{The figures above show the first two dimensions of each difficulty of overlap problem. As the problem difficulty increases, the overlap between the two components increase. Therefore, in the highest difficulty setting, the points are poorly separated which leads to large amounts of overfitting. It is in settings such as these that we believe the Thresholding Method will provide the greatest benefit.}
\end{figure}
\vspace{4mm} \noindent After determining the Misclassification Rate and Probability of Assignment for each value of $t$ we apply the Thresholding Method on the fitted and true likelihoods. By applying the Thresholding Method we demonstrate the effectiveness of the method in providing control over the misclassification rate for each difficulty problem.

\vspace{4mm} \noindent Figure 2 shows the results of the thresholding method with target $q_{easy} = .25$, $q_{medium} = .25$, and $q_{hard} = .25$ . The points on Figure 2 on each fitted likelihood plots indicates the $t^*$ that should be selected, as well as the number of points that should be classified in the testing set to ensure a misclassification rate below the target.

\vspace{4mm} \noindent Below we show the first two dimensions of the mixture model for each difficulty. Using the simulated data, we compare the Misclassification Rate to the Probability of Assignment for two likelihood schemes. In the fitted likelihood case, we estimate the sample means ($\hat{\mu}_1$, $\hat{\mu}_2$), covariance matrices ($\hat{\Sigma}_1$,$\hat{\Sigma}_2$), and mixture components ($\hat{\pi}_1, \hat{\pi}_2$), from the data and use the fitted parameters to calculate Misclassification Rate and Probability of Assignment for each $t$ on each dataset. In the second, true likelihood case, we use the oracle values for the components to calculate the Probability of Assignment and Misclassification Rate for all three datasets. Further, as the hold-out and testing datasets are individual subsamples from the dataset, by repeatedly drawing new testing and hold-out sets we are able to average the lines over several simulations to determine the expected Misclassification Rates and Probability of Assignment for each $t$ for both the fitted and true likelihood cases.

\vspace{4mm} \noindent As seen in Figure 2, all three difficulty problems show significant overfitting on the fitted likelihoods. Therefore, due to the overfitting, it may be difficult to use the training data to determine which points should be classified in order to control the misclassification rate. As the difficulty of the problem increases, the misclassification rates increase due to the more pronounced overfitting and challenge in fitting non-separable data. 

\vspace{4mm} \noindent The difficulty of the problem also can be seen in the true likelihoods. As the overlap increases, the training, testing, and hold-out errors increase. Generally as the difficulty increases, the proportion of points assigned to achieve a low error rate decreases. This is because in the less separable settings, more of the points lie between the two mixture components and are difficult to classify. However, in all three cases, the training error on the fitted likelihood was near or at zero indicating large amounts of overfitting.

\vspace{4mm} \noindent In these cases, the training data alone would not be very informative in order to gauge the error on the testing datasets. In the hardest problem setting especially, if all points were assigned as determined by suggested by the training errors, on the testing error there would be an unsatisfactory testing error of 26\%. 

\vspace{4mm} \noindent The use of the hold-out set in the Thresholding Method provides an avenue to assess the magnitude of overfitting within the classification scheme. When heavy overfitting the present, few points are assigned as the hold-out set effectively provides the conservative $t^*$ needed to sift out the strongest signals from the data. The improvement over the naive method through the use of this hold-out data will be explored later in this section.

\vspace{4mm} \noindent Further, this problem was generated in 30 dimensions. As will be explored in the next simulation study, the problem of overfitting is amplified by increasing the dimensionality as well. In a very high-dimensional setting with moderate overlap, the effect of overfitting will become more pronounced. Therefore, the training data alone will provide very little information about the errors in the testing set.
\begin{figure}[H]
\begin{centering}
{\includegraphics[scale=0.31]{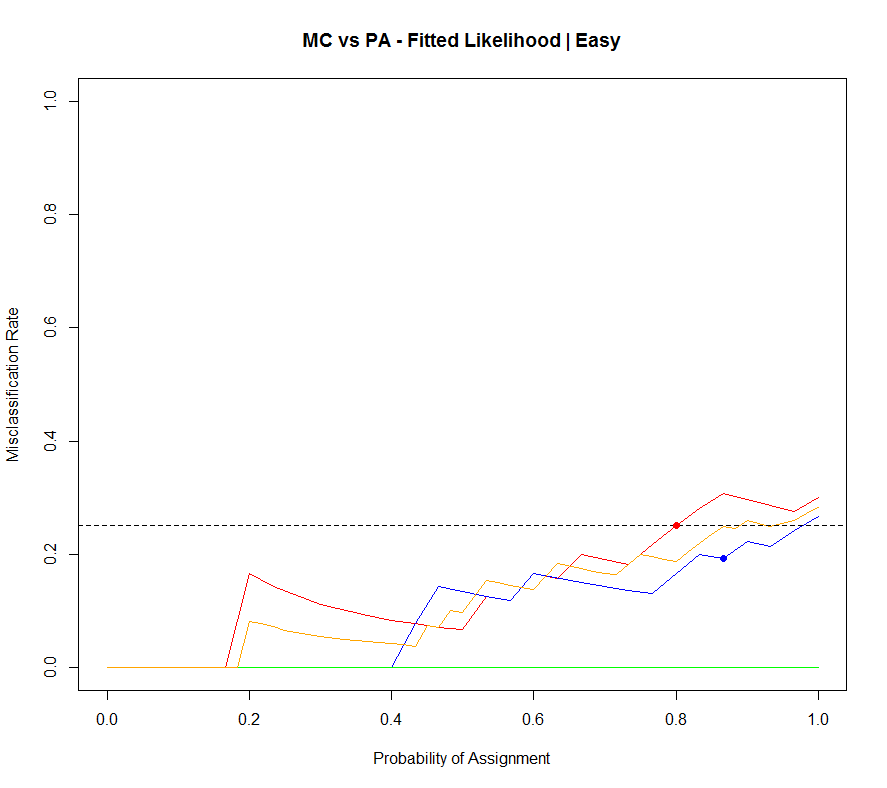}}{\includegraphics[scale=0.31]{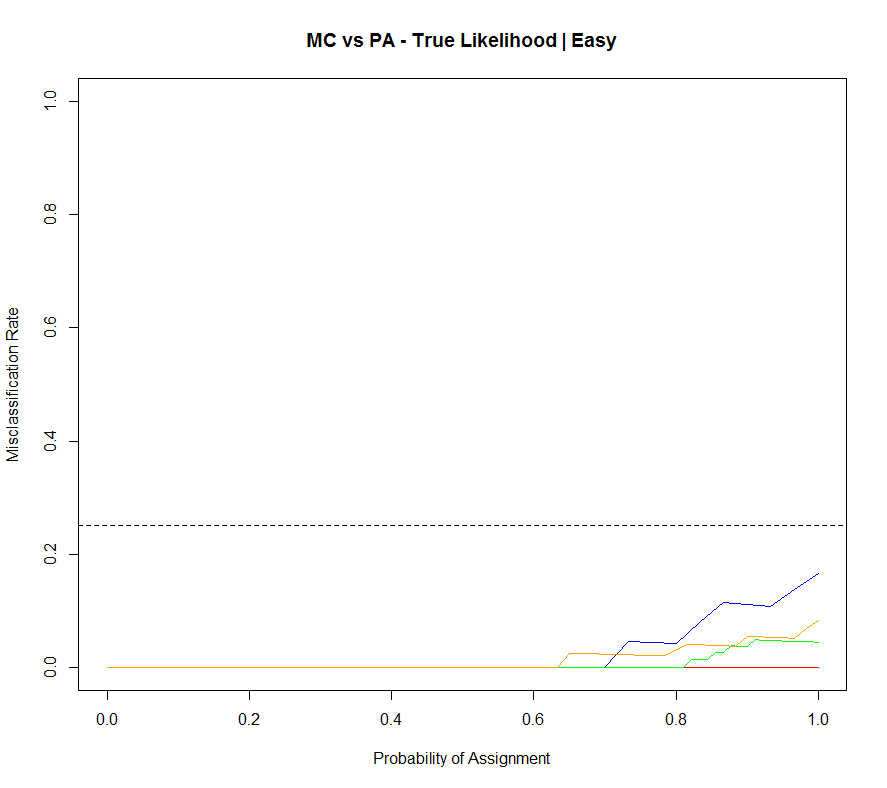}}
{\includegraphics[scale=0.31]{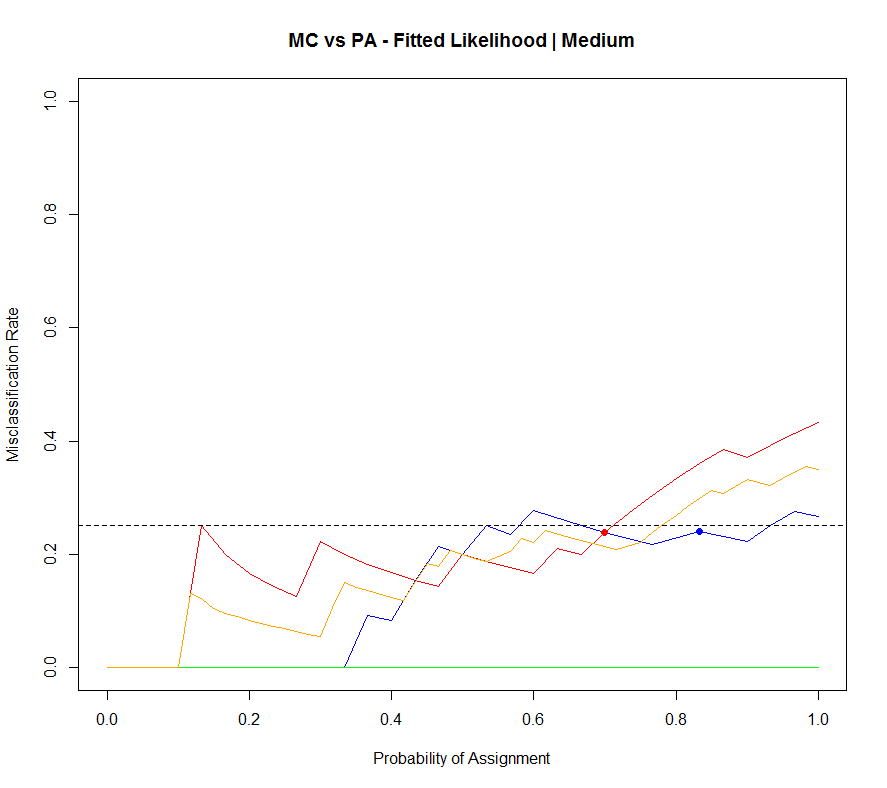}}{\includegraphics[scale=0.31]{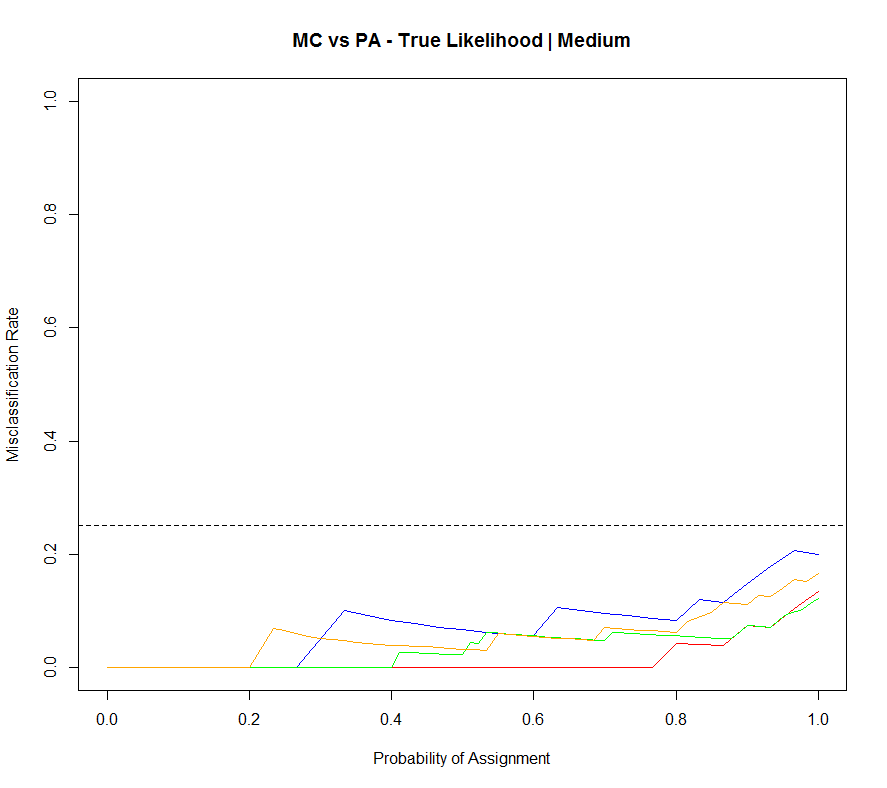}}
{\includegraphics[scale=0.31]{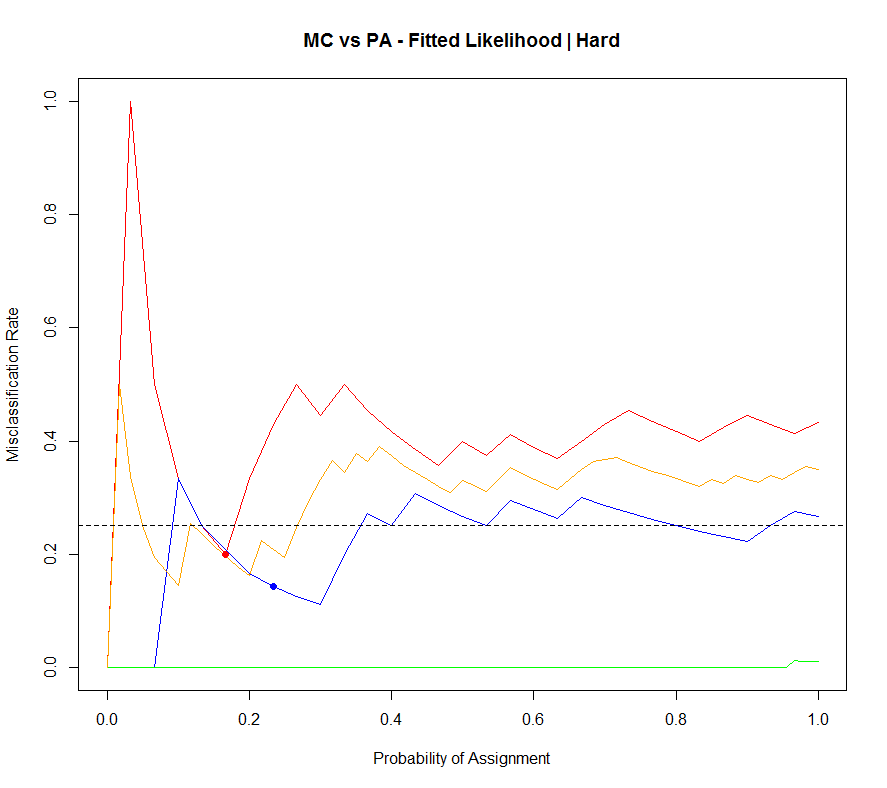}}{\includegraphics[scale=0.31]{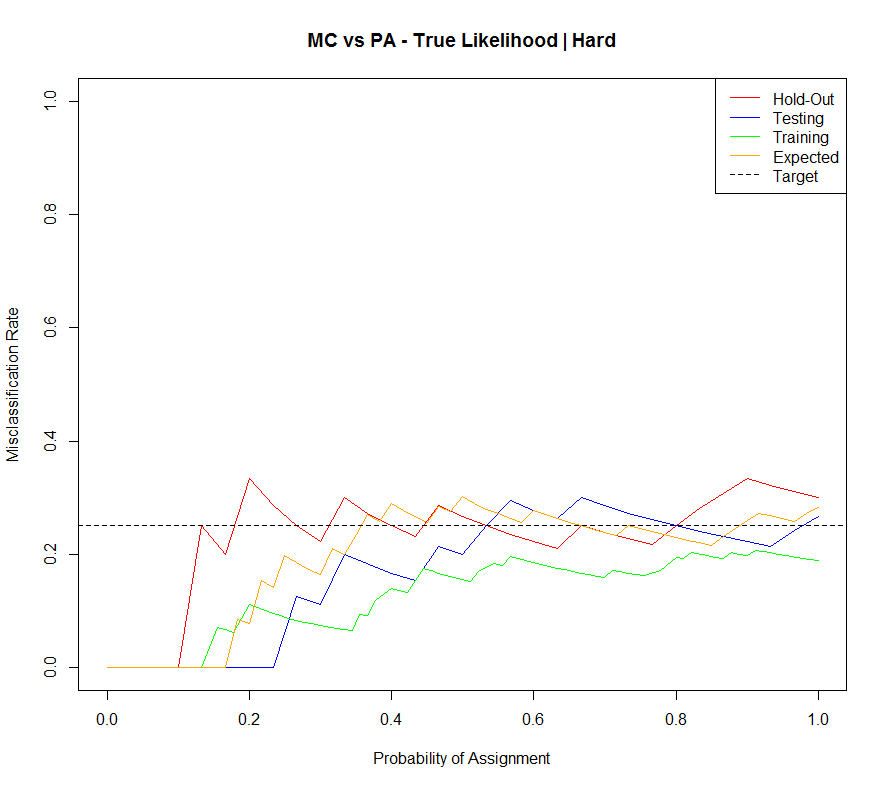}}
\par\end{centering}
\caption{MC vs PA for the 2 component Gaussian mixtures where difficulty is controlled by the distance of the Gaussian centers. As the difficulty of the problem increases, the centers are brought closer to one another and the effect of overfitting increases. The points on the plots indicate the associated MA and PA on the testing and hold-out sets respectively when the Thresholding Method at $t^*$ is run.}
\end{figure}

\vspace{4mm} \noindent As seen in the hold-out data, the curve indicates large amounts of overfitting occurred in the hard problem case. Therefore, by then applying the Thresholding Method, we are able to have better control of our expected error in the testing set. In the hard case, by setting the cutoff at 25\%, the method yields the associated $t^*$ of 19.57 on the hold-out set, corresponding to a hold-out error of 20\%. When this $t^*$ is used to classify the testing data, this yielded a testing error of 14\% with 23\% of the points classified. This error is satisfactory as it is below our cutoff. This process can be repeated for both the medium and easy problems themselves and the results are summarized in Table 1. 

\begin{table}[H]
\begin{centering}
\begin{tabular}{|c|c|c|c|c|c|}
\hline 
\textit{Dataset} & \textit{Target q} & \textit{Hold-Out MC} &\textit{Hold-Out PA} & \textit{Test MC} & \textit{Test PA} \tabularnewline
\hline 
\textit{Easy} & .25 & .25 & .80 & .19 &.87\tabularnewline
\textit{Medium} & .25 & .24 & .70 & .24 & .83\tabularnewline
\textit{Hard} & .25 & .20 & .17 & .14 &.23\tabularnewline

\hline 
\end{tabular}
\par\end{centering}
\caption{MC and PA on Hold-out and Testing sets for each center position difficulty after Thresholding Method is applied at each target $q$.}
\end{table}
\noindent The use of the hold-out set allows for a gauging of the overfitting. By parameterizing the problem by $t$ we are able to exert control over a desired classification error. By knowing the errors across $t$ in the hold-out set, we are then able to have control over what errors we would expect to see when classifying a new dataset from the same population. As the overlap between components increases, more points fall in the margin where it is difficult to classify the point. Therefore, to achieve low error rates very few of the points, only those with strong signals, should be classified.

\subsubsection{Difficulty from Dimensionality}

\noindent In the next setting, we control the difficulty of the problem via dimensionality. We expect that as the dimensionality of the problem increases, the difficulty in estimating the underlying GMM and classifying the points increases. We show that the Thresholding Method provides control over the error rate on the testing set even as the difficulty of the underlying problem changes.

\vspace{4mm} \noindent Even as $d$ increases, for purposes of comparing across dimension, we ensure that the \textit{Total Variation Distance} [1] remains constant. 

\vspace{4mm} \noindent \textit{Total Variation Distance (TVD) between densities f(x) and g(x)}:

$$TVD = ||f(x)-g(x)||_{TV} = \frac{1}{2} \int | f(x) - g(x) | dx $$

\vspace{4mm} \noindent For efficient computation time, instead of directly calculating this integral we sample repeatedly and estimate the total variation distance through:

$$TVD = \mathbb{E}_g[|\frac{f(x)}{g(x)}-1|]$$

\vspace{4mm} \noindent Therefore, the problems remain appropriately comparable across dimension. This ensures the points are not too spread in high-dimensional cases and so the likelihoods are not too large in these large dimension settings. While this renders the comparison across dimensions fair for classification, the increase in the dimensionality still affects the performance of the classification and estimation schemes.

\vspace{4mm} \noindent For the three cases, $n=500$. Further, the centers were fixed across all three problems, once again generated a two-dimensional problem embedded in the larger $d$-dimensional space. Two component mixtures were used again with equal mixture proportions. The true covariance matrices were diagonal, though the elements of the diagonal were changed across dimension to control the TVD. Three different dimensionalities were selected with $d \in \{10, 20, 100\}$. As such, to ensure constant TVD across dimensionality, the diagonal entries of $\Sigma_1$ and $\Sigma_2$ were set to [$\frac{8}{10}$,$\dots$,$\frac{8}{10}$]$^{10}$, [$\frac{3}{4}$,$\dots$,$\frac{3}{4}$]$^{20}$, and [$\frac{1}{2}$,$\dots$,$\frac{1}{2}$]$^{100}$ respectively. The process of fitting the GMM and running the Thresholding method proceeded identically to the previous simulation setting with a target $q = .2$ for all problem difficulties. The misclassification rates and proportion of points assigned for each difficulty problem for both testing and hold-out data are shown in Table 2 and the plots are seen in Figure 3.

\begin{figure}[H]
\begin{centering}
{\includegraphics[scale=0.32]{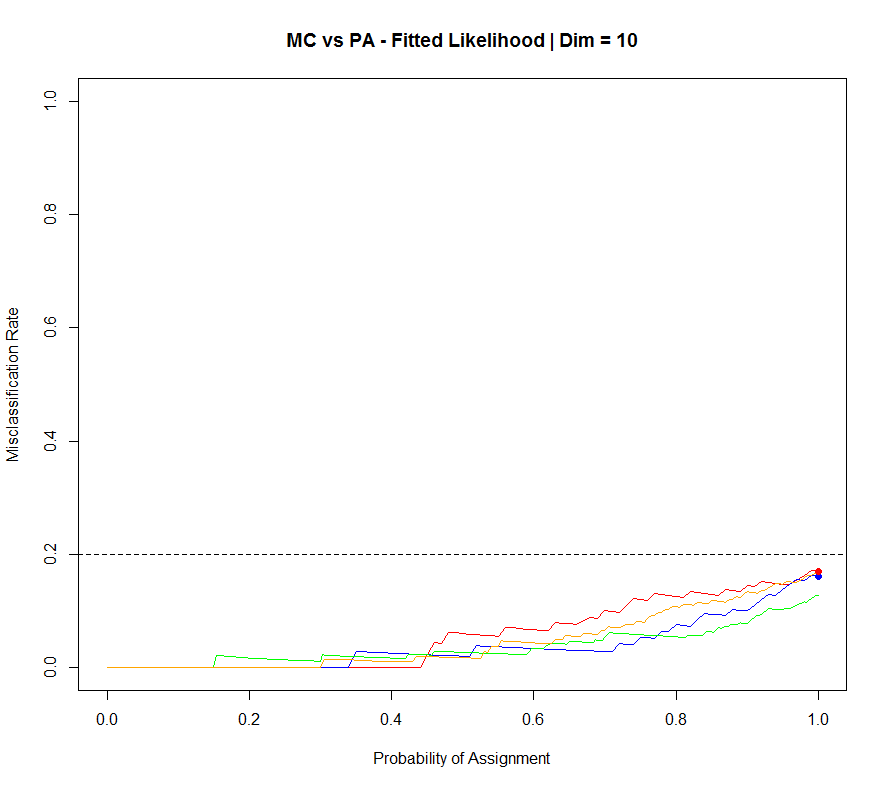}}{\includegraphics[scale=0.32]{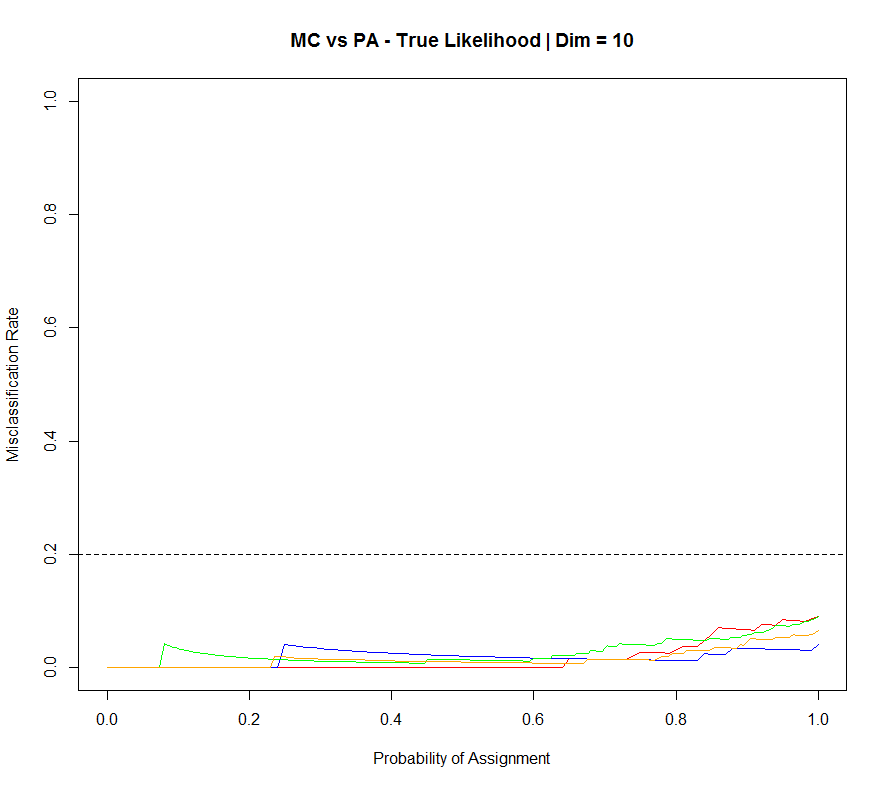}}
{\includegraphics[scale=0.32]{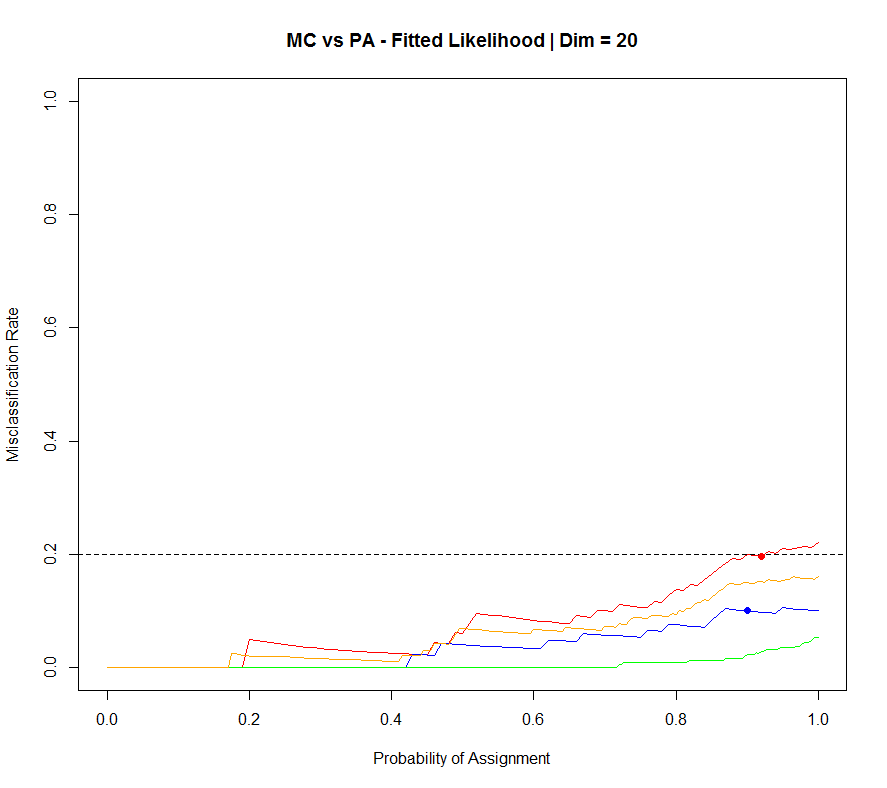}}{\includegraphics[scale=0.32]{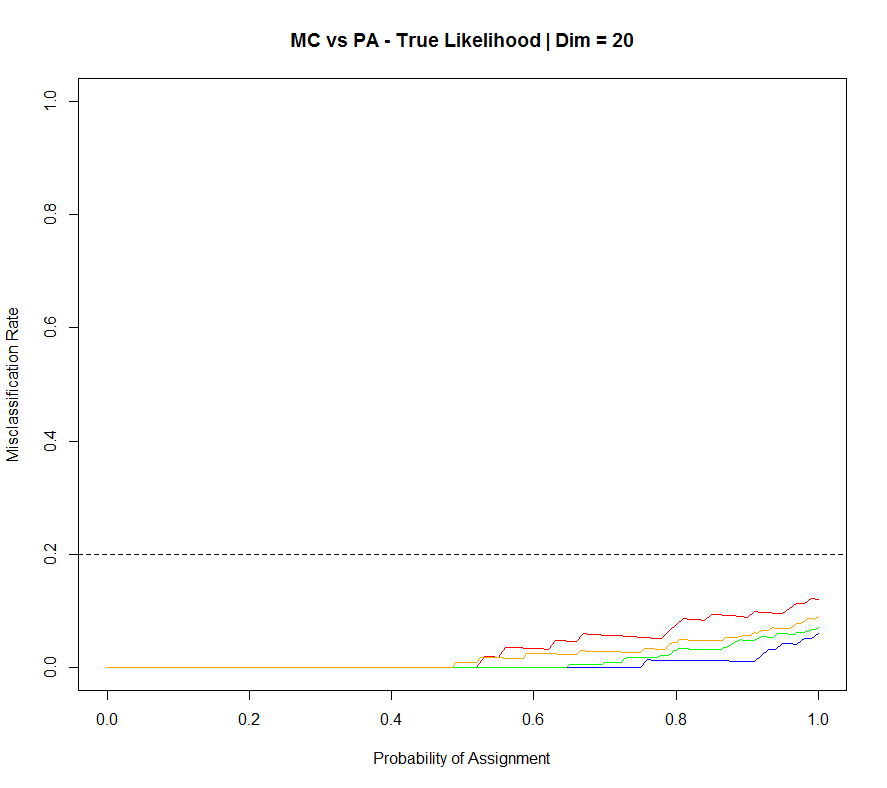}}
\par\end{centering}
\end{figure}
\begin{figure}[H]
\begin{centering}
{\includegraphics[scale=0.32]{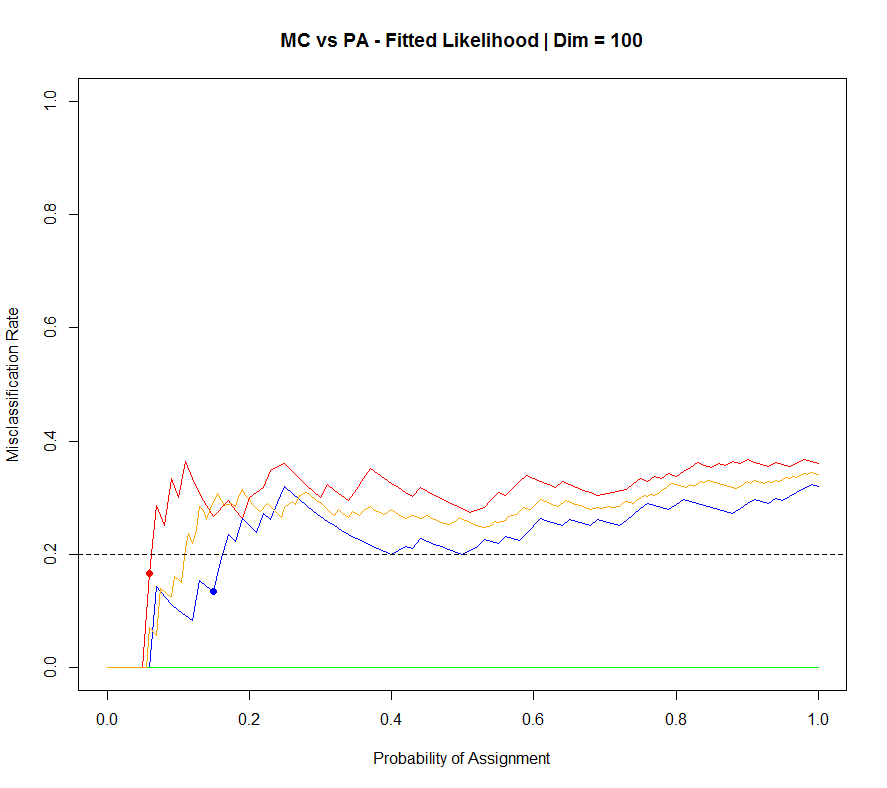}}{\includegraphics[scale=0.32]{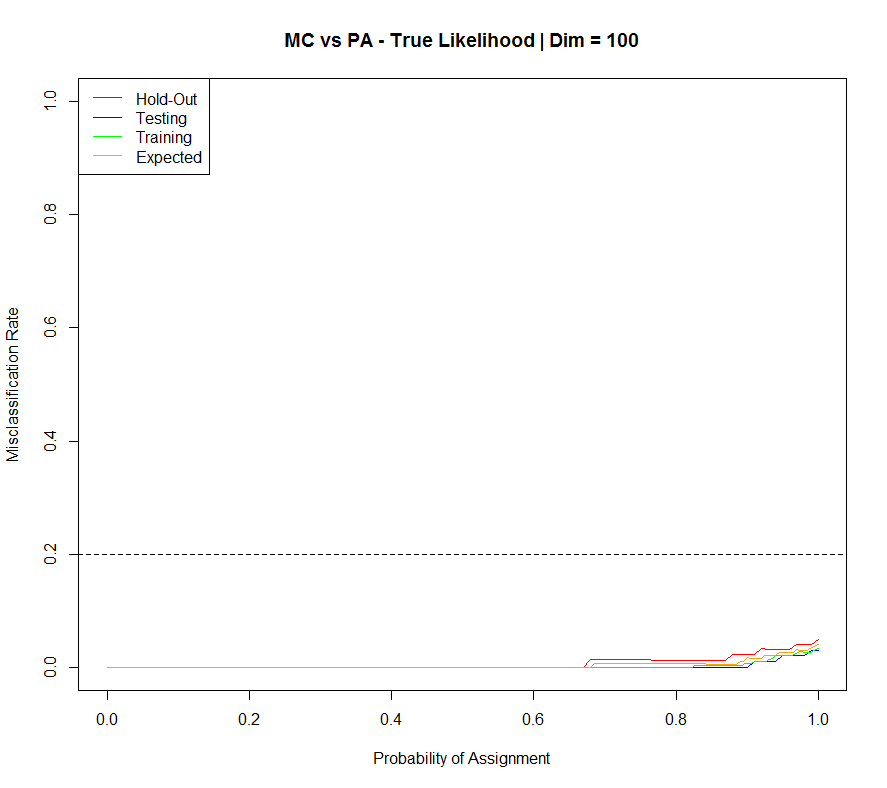}}
\par\end{centering}
\caption{MC vs PA for the 2 component Gaussian mixtures where difficulty is controlled by the underlying dimensionality of the problem. As the dimensionality increases, the difficulty of the problem increases as the magnitude of overfitting increases. The points on the plot are the associated MA and PA on the testing and hold-out sets respectively when the Thresholding Method at $t^*$ is run.}
\end{figure}

\noindent Firstly, in all three scenarios, the Thresholding Method provides control over the misclassification error from the training set. While in the lower dimensional cases, the naive method of selecting the classified points through the training error would work sufficiently, in the highest-dimensional setting this would be a poor decision. In the most difficult scenario, there was no training error in the fit; however, large amounts of overfitting occurred. Without using the hold-out error to gauge the amount of overfitting, if all points were classified, the error would be more comparable to 33\%, which is above the target. As such, the Thresholding Method provides assistance in validation in the most difficult cases by providing a gauge for the magnitude of overfitting that has occurred.

\begin{table}[H]
\begin{centering}
\begin{tabular}{|c|c|c|c|c|c|}
\hline 
\textit{Dataset} & \textit{Target q} & \textit{Hold-Out MC} &\textit{Hold-Out PA} & \textit{Test MC} & \textit{Test PA} \tabularnewline
\hline 
\textit{Easy} & .2 & .17 & 1.00 & .16 & 1.00\tabularnewline
\textit{Medium} & .2 & .20 & .92 & .10 &.90\tabularnewline
\textit{Hard} & .2 & .17 & .06 & .13 & .15\tabularnewline
\hline 
\end{tabular}
\par\end{centering}
\caption{MC and PA on Hold-out and Testing sets for each dimensionality difficulty after Thresholding Method is applied at each target $q$.}
\end{table}

\noindent As the dimensionality of the problem increases, the Thresholding Method tends to be more conservative and classifies fewer of the total points. This conservative approach prevents points near the decision boundary from being classified. As the higher-dimensional settings yielded more overfitting, the Thresholding Method corrects for this occurrence by being stringent and only classifying the points with the larger log-likelihood differences. While, for example, in the hardest difficulty setting this yielded only 15\% of the points being classified, the error was successfully controlled at 13\%. However, if the problem setting allows for higher possible error, more points would be classified by relaxing the $q$ selection to a less conservative level.

\subsection{Thresholding Method and Naive Method Comparison}
\noindent Next, we compare the effectiveness of the Thresholding Method on simulated data and the naive method. For fairness, as the quality of the fit is dependent on the number of training samples, we compare the naive method where the training data consists of the training and hold-out set points, or equivalently, combining the training and hold-out into one large dataset for fitting The Thresholding Method uses a hold-out set, thereby having fewer points to train the model with. The testing sets remained the same between the two methods.

\vspace{4mm} \noindent In the naive method, there is no hold-out set to gauge the overfitting or model misspecification. Instead, the additional data points are used to improve the accuracy of the underlying fit. The Thresholding Method uses fewer points for the fit, but instead uses the hold-out points to better estimate which points should and should not be classified in order achieve a target misclassification rate.

\vspace{4mm} \noindent In Figure 4, a single sampling of training, testing, and hold-out points was performed on the medium difficulty centers problem from the earlier simulation study at a lower dimensionality of 15 dimensions. In the naive method, 60 points were used to fit the GMM and the 15 point testing set was evaluated at each thresholding point $t$ using the fitted GMM. Likewise, in the method with hold-out data, the 45 point training dataset was used to estimate the GMM and 15 points were used to construct the validation set with the same 15 point testing between the two settings.

\vspace{4mm} \noindent When the hold-out set is used, we have more control over the error due to having an estimate of what the testing error is expected to be. By viewing the hold-out set curve, this indicates that there was large overfitting in the data. Through the curve, we are able to determine whether all points should be classified or not in order to achieve a desired target. This is not possible with the training data alone because all the errors for all $t$ are below the threshold. Therefore, this indicates that we may classify all the points because we have know knowledge of the overfitting, which fails to control the error in the testing set. While the training error in the scenario without the hold-out set was lower, due to having more points to fit the model, we sacrifice the ability to validate the model for overfitting.

\vspace{4mm} \noindent We now repeat this process multiple times, as errors are dependent on the initial subset for the training, hold-out and testing data. We resample 50 times and repeat the previous method. By averaging across the resampled data, we demonstrate the effectiveness of the validation set in the Thresholding Method over the naive method in gauging and controlling for the overfitting in the data. Table 3 shows the difference between the methods when attempting to control at the $q = .2$ level after applying the Thresholding Method and naive methods respectively. The Avg. Test MC refers to the average testing Misclassification Rate after using the Thresholding Method to determine the target $t^*$ using the hold-out and training errors respectively. Avg. Test PA is calculated in a similar manner.

\begin{figure}[H]
\begin{centering}
{\includegraphics[scale=0.35]{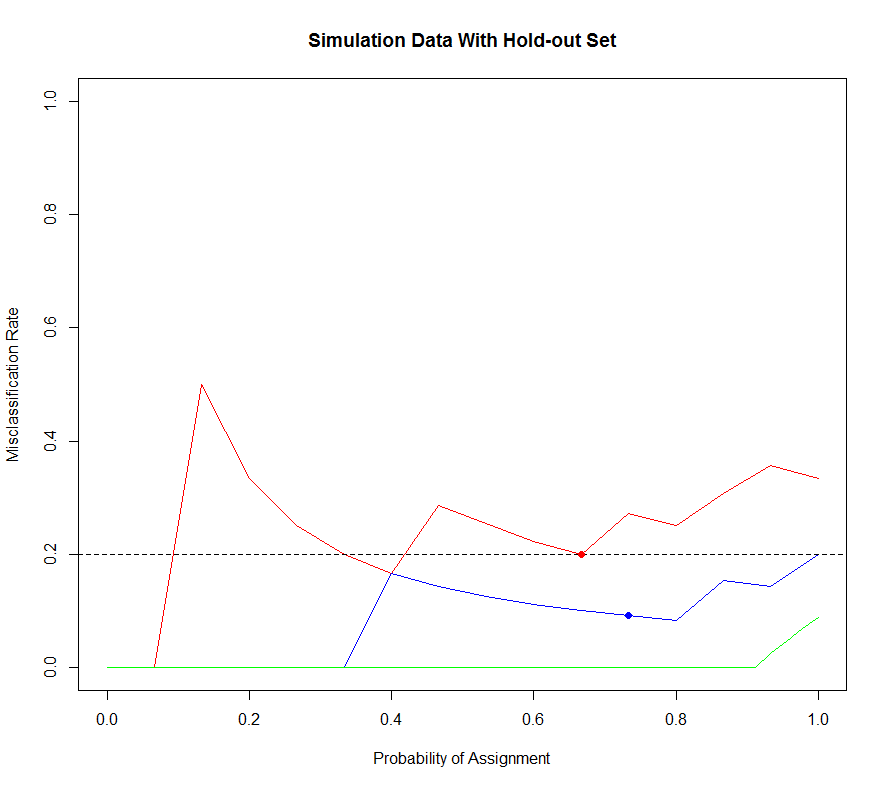}}{\includegraphics[scale=0.35]{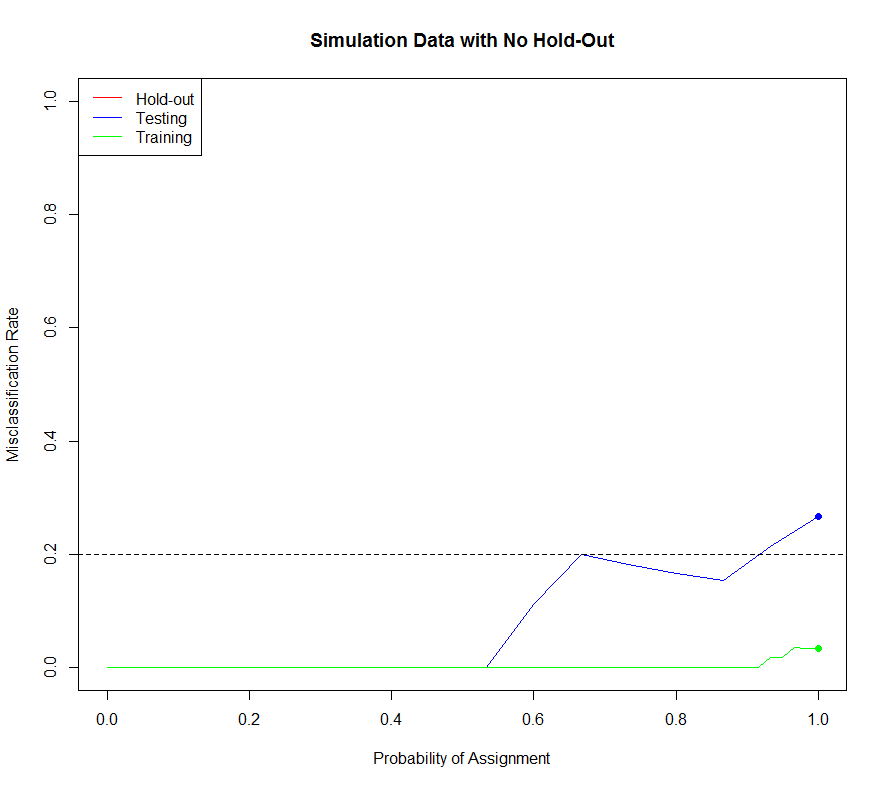}}
\par\end{centering}
\caption{MC vs PA for two cases from a single instance. The first is through running the Thresholding Method with a hold-out set as a sample of the training points thereby using a smaller training set to fit the GMM. The second combines the hold-out and training set to create a larger training set used to fit GMM. The points on the points are the associated MC and PA determined by applying the Thresholding Method and naive method at the respective $t^*$ values.}
\end{figure}
\begin{table}[H]
\begin{centering}
\begin{tabular}{|c|c|c|c|}
\hline 
\textit{Dataset} & \textit{Target q} & \textit{Avg. Test MC} &\textit{Avg. Test PA} \tabularnewline
\hline 
\textit{With Validation Set} & .2 & .152 & .069 \tabularnewline
\textit{Without Validation Set} & .2 & .312 & 1.00 \tabularnewline
\hline 
\end{tabular}
\par\end{centering}
\caption{MC and PA for Thresholding Method using the validation set and the naive method without validation set with a target $q = .2$. The results are the MC and PA averaged over 50 resampled trials}
\end{table}
\noindent The naive method to control at the .2 level indicates that all the data points should be classified. However, this yielded a misclassification rate of .312. Without the hold-out set, there is no knowledge of the amount of overfitting until the testing set is classified. Therefore, the control is poor without the validation set. When the validation set is used, only 7\% of the testing set tends to be classified on average. However, this also yields a satisfactory misclassification rate control of .152 which is below the target $q$.

\vspace{4mm} \noindent Further, the targeted $q$ in this simulation was selected to show the effect of overfitting in control without the hold-out set. The hold-out curve also provides the benefit of showing the magnitude of overfitting in the data without having to fit the testing set. Therefore, by knowing the magnitude of overfitting, the $q$ could have been selected to be higher such as at $q = .45 $ to account for the overfitting and more points would be classified. Without the hold-out set and only the training error curve, little control is provided as the training errors are low.

\vspace{4mm} \noindent The use of the hold-out set provides calibration not only to correct overfitting, but also any model misspecification. In higher-dimensional settings, it is often difficult to determine if the model we choose to fit is indicative of the underlying data distribution. Without the use of the hold-out set, we may have fit a model with a low training error, but the high testing error produced may be due to a poor model choice. The Thresholding Method provides a means of calibration in this scenario, by using the hold-out set to adapt to poor model fits. If a model is heavily misclassified, this would often be revealed through high error on the hold-out set, which would yield a small subsection of the data being classified to achieve a low error threshold.

\section{Real Data Results}
\noindent In this section, we apply the Thresholding Method to two different real datasets. Firstly, we will apply the Misclassification Rate vs Probability of Assignment scheme and Misclassification Loss vs Average Entropy scheme to the ionosphere data set [4] provided through the UCI Machine Learning Repository [2]. After, we demonstrate the flexibility of the Thresholding Method by applying the Misclassification Rate vs Probability of Assignment scheme to two different classification methods using the ozone level data set [3] from the UCI Machine Learning Repository.

\subsection{Ionosphere Data}
\noindent The ionosphere dataset is a binary classification dataset. The features in question are readings from radar data consisting of an array of high-frequency antennas. Using these collected radar responses, the signal was then classified as a \textit{good} or \textit{bad} signal. A \textit{good} signal is one that exhibited a certain structure in the ionosphere while the \textit{bad} signal did not show that structure.

\vspace{4mm} \noindent In this dataset, the features are 34 pulses from the radar system and the response is a binary variable for \textit{good} and \textit{bad} radar returns. The total number of samples in this dataset is 351. In order to construct, validate, and test our classifier, the dataset was split into 3 sets. The training dataset consisted of 151 observations and the validation and testing sets contained 100 observations each.

\vspace{4mm} \noindent Firstly, using the labeled training data, a mixture model was fit, estimating $\hat\mu_{bad}$, $\hat\mu_{good}$, $\hat\Sigma_{good}$, $\hat\Sigma_{bad}$, $\hat\pi_{good}$, $\hat\pi_{bad}$. Using these parameters, the Thresholding Method using the Misclassification Rate Formulation was performed. In running this model, the enumerated grid of \textit{T} was constructed in order to maximize computation efficiency as well as properly catch when a point leaves the set of classified points. Using the full dataset, the likelihood of each point for each class given the training estimates for population parameters was calculated. As this is a binary classification scenario, the absolute difference of the two likelihoods was calculated and sorted to enumerate the grid of \textit{T}. Therefore, each value of \textit{T} is the point where an element of the dataset leaves the classification set. $\infty$ and a small $\epsilon$ were appended to the training grid for the settings where nothing is classified, and all points are classified respectively.

\vspace{4mm} \noindent By applying the Thresholding Method on the training dataset, the curve for the Misclassification Rate vs Probability of Assignment was calculated. As the training dataset was also used to fit the model, a large amount of overfitting can be seen in comparison to the hold-out and testing sets. 

\vspace{4mm} \noindent The Thresholding Method using the Misclassification Rate also yields a curve for the misclassification rate vs probability of assignment for the testing set. By plotting these values, it once again reveals significant overfitting. A target $q = .15$ was selected in order demonstrate the method. Therefore, we want to determine the thresholding $t^*$ such that the Misclassification Rate on our testing data will be below 15\%.

\vspace{4mm} \noindent Secondly, the Thresholding Method using the Misclassification Loss was used on the same training, testing, and validation set. At each point of the same grid $T$, the Misclassification Loss and Average Entropy were calculated for the training, validation, and testing datasets. In this example, the maximum classification loss tolerated, $r$, was set to $r = 300$. By determining the minimum $t^*$ such that the Misclassification Loss on the hold-out data was below $r$, we then use this threshold to classify the testing data so that the Misclassification Loss on the testing data is below our desired limit.

\begin{figure}[H]
\begin{centering}
{\includegraphics[scale=0.36]{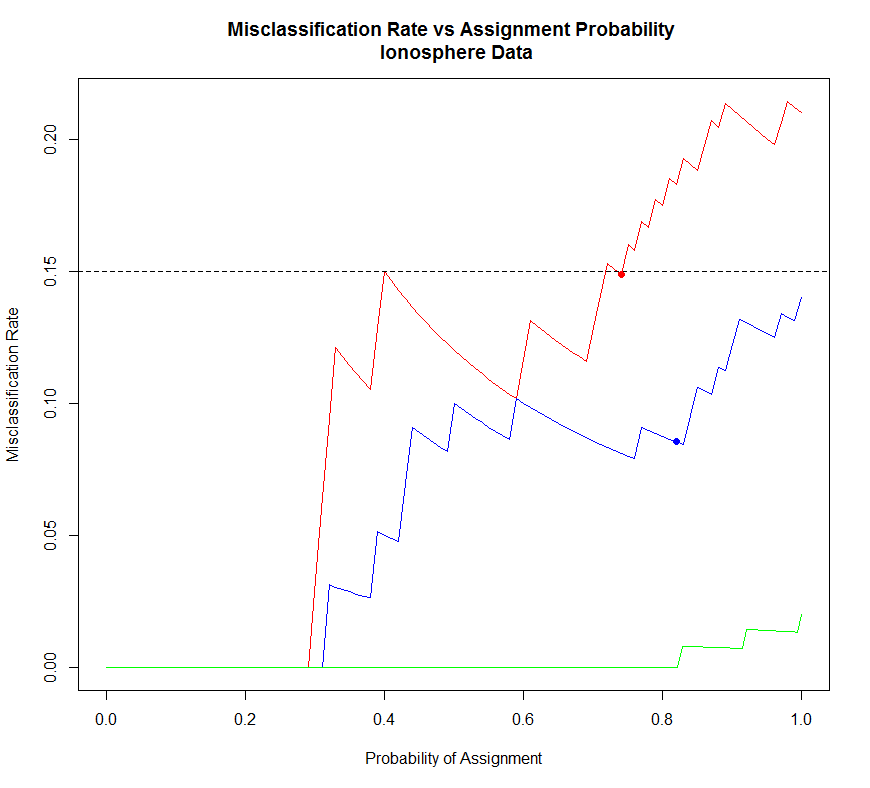}}{\includegraphics[scale=0.36]{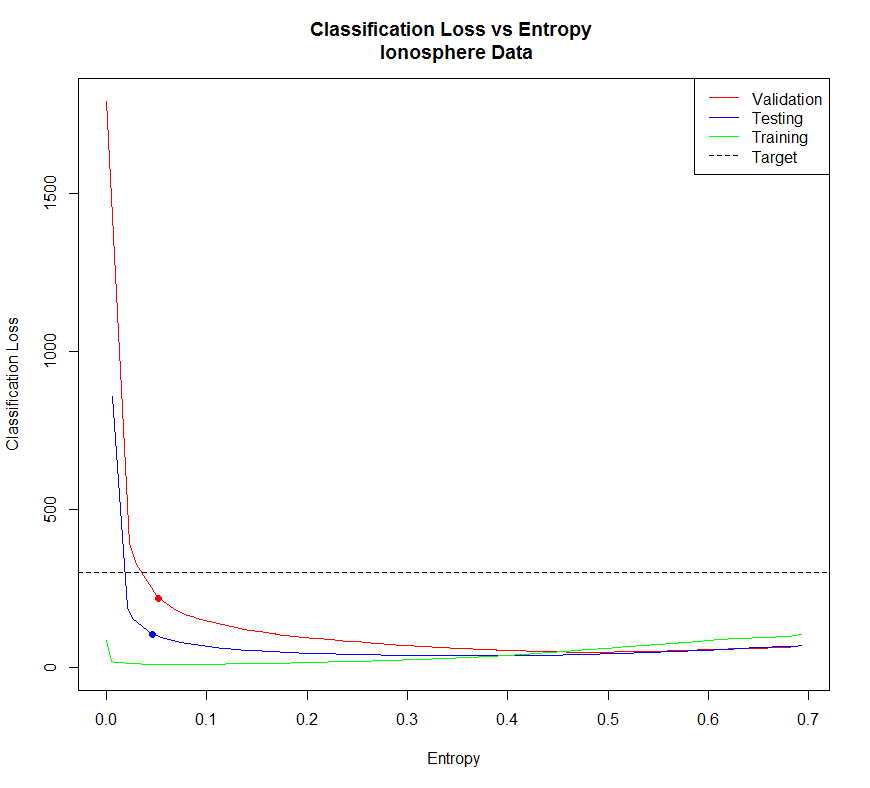}}
\par\end{centering}
\caption{The application of the Thresholding Method to the ionosphere data using both the Misclassification Rate and Misclassification Loss formulations. The data was split into training, testing, and hold-out sets and desired target $q =.15$ for the misclassification rate version and $r=300$ for the classification loss formulation. The points on the plot represent the MC and PA, and MCL and AE respectively from classifying using the $t^*$ choices determined by the Thresholding Method for each given loss function.}
\end{figure}

\noindent In both the Figure 5 plots, we see evidence of overfitting to the training data. The Misclassification Rate is much lower on the training data than the validation and testing data. This is a common scenario in data analysis where the initial choice of method fits closely to the training data. Using the validation data, we are able to view this overfitting and appropriately choose $t^*$ to control the error rate. By setting the threshold to $q= .15$, to achieve this target Misclassification Rate, we must assign roughly 74\% of the points in the hold-out set at a hold-out Misclassification Rate of 15\%. The 26\% of the points that are not classified are the points where it is difficult to ascertain the label and lie nearer to the decision boundary. When applied to the testing data, this yielded a Misclassification Rate of approximately 8\% and assigned 82\% of the points, which is below our desired bound.

\vspace{4mm} \noindent A similar scenario occurs in the Misclassification Loss formulation plot. In this case, we once again see overfitting in the testing data, especially when the entropy is low. The three lines quickly converge near each other as we increase $t$. Therefore, as $t$ increases, a more conservative approach is taken as we wish to only classify points with class assignments we are certain about. Once again in this example, by setting a Misclassification Loss target at $r = 300$, on the validation set, we select the $t^*$ that most closely yields a Misclassification Loss near the target. This associates to a threshold that yielded a Misclassification Loss of 218 with an Average Entropy of .052 on the dataset. By using this $t^*$ on the testing set, we once again see that the testing Misclassificaton Loss was below the threshold at 105 with an Average Entropy of .046.

\vspace{4mm} \noindent Therefore, by parameterizing the problem with $t$, we are afforded control over our classification scheme through the appropriate selection of $t$. The Thresholding Method allows us to only classify the points deemed the most informative, given a maximum tolerance on our selected loss function. As demonstrated, the Thresholding Method can be adapted to different loss functions lending further flexibility.

\subsection{Ozone Data}
\noindent The ozone dataset is a binary classification dataset as well. The features in this dataset are a mix of ozone levels, wind speeds, radiation values, temperatures and other atmospheric readings. The classification scheme are \textit{ozone days} and \textit{non-ozone days}.

\vspace{4mm} \noindent There are 72 features within this dataset, with 2536 observations. As many of the observations contain missing data, those observations with a missing value were omitted from the dataset. Therefore, the number of observations was 1848. This dataset was selected to show the flexibility of Thresholding Method. The samples are extremely imbalanced in this dataset with only 57 of the readings being classified as \textit{ozone days}, therefore the mixture probabilities are heavily skewed towards \textit{non-ozone days}.

\vspace{4mm} \noindent We demonstrate the Thresholding Method on two versions of the classification problem. In the first setting we use the full dataset, where all features and data points were used. Due to the imbalanced mixture proportions, this may not be ideal as the data is skewed towards \textit{non-ozone days}, affecting the classification quality. The second version is one where the data was subsampled in order to more evenly split \textit{ozone days} and \textit{non-ozone days} in the training dataset. In this version the 57 \textit{ozone days} data points were sampled in the training set. Next, 143 \textit{normal days} were sampled randomly from the dataset, creating a new dataset with 200 observations and less-skewed mixture proportions. Further, due to the smaller dataset, the top 20 features with the highest marginal correlation to the training labels were chosen. 

\vspace{4mm} \noindent For the full dataset version, the training, testing, and validation set data each consisted of 616 observations each. For the subsampled data, the training set consisted of 100 observations, while the testing and validation set contained 50 observations.

\vspace{4mm} \noindent Using both these versions, the full and reduced with marginal correlation feature screening, the mixture model was fit estimating $\hat\mu_{oz}$, $\hat\mu_{no}$, $\hat\Sigma_{oz}$, $\hat\Sigma_{no}$, $\hat\pi_{oz}$, $\hat\pi_{no}$ using the training data. The Thresholding Method using the Misclassification Rate Formulation was applied to the training, validation, and testing datasets and the curves are shown in Figure 6.

\begin{figure}[H]
\begin{centering}
{\includegraphics[scale=0.36]{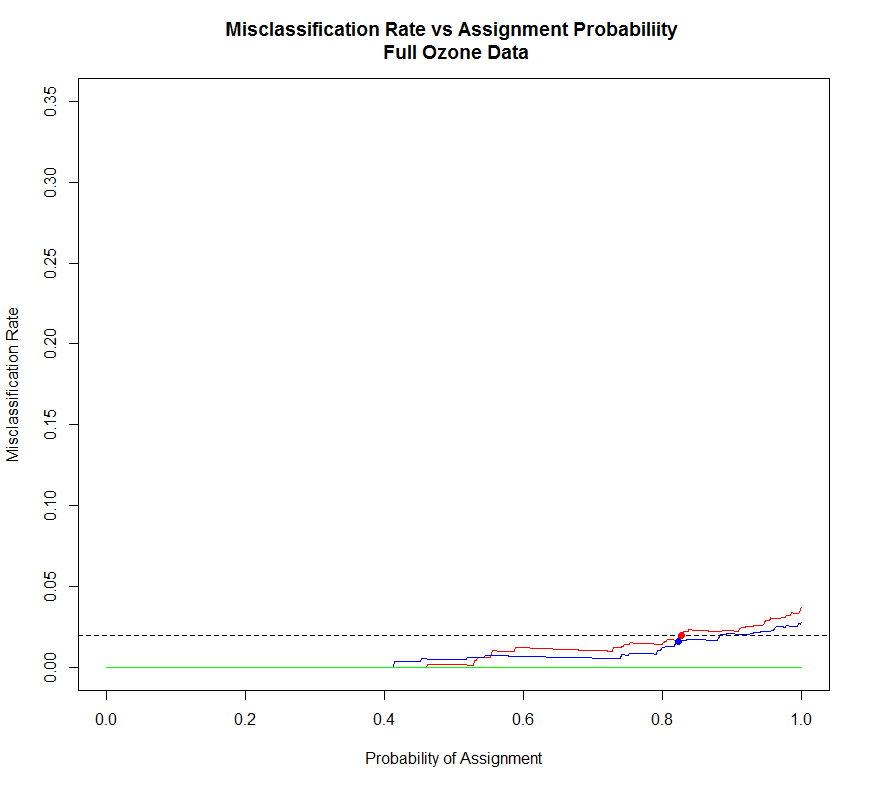}}{\includegraphics[scale=0.36]{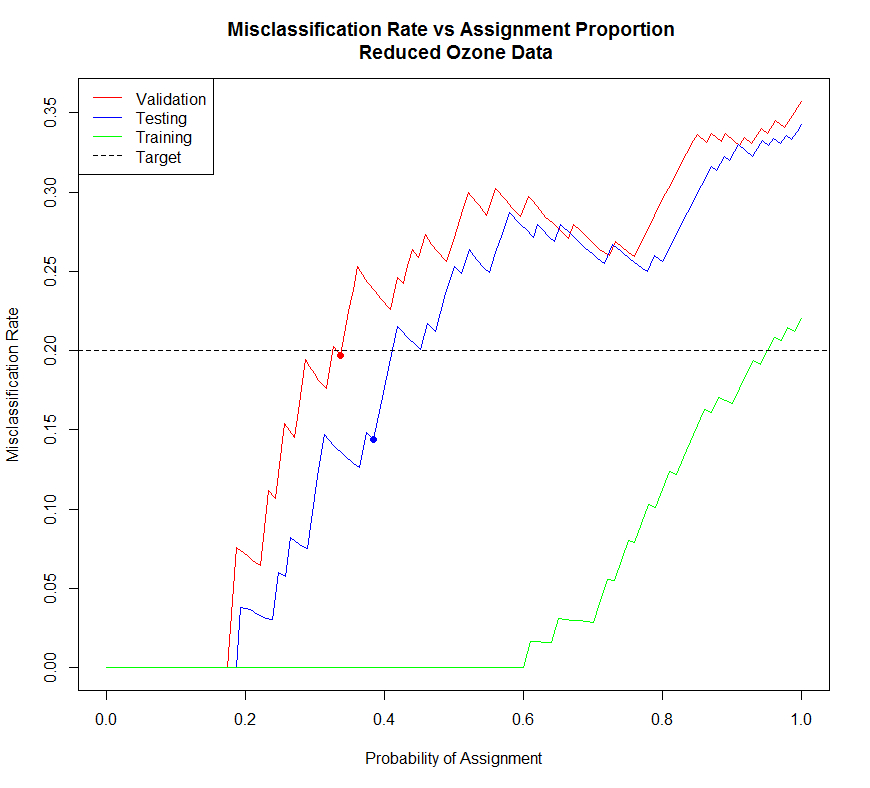}}
\par\end{centering}
\caption{The application of the Thresholding Method to the ozone data. In the full ozone data, all of the data was used and split into training, testing, and hold-out sets. In the reduced ozone data, the data was subsampled to more evenly balance the number of \textit{ozone days} and \textit{non-ozone days} points in the training data. Further, in the reduced ozone data, the feature space was reduced to only include the 20 features with the highest marginal correlation to the labels. The points on the plot are the MC and PA values for the Thresholding Method being applied at the $t^*$ values for each of the models.}
\end{figure}

\noindent In the full dataset version, a target misclassification rate of $q = .02$ was chosen. By using the validation data curve, the $t^*$ was selected to maximize the probability of assignment while still ensuring a target Misclassification Rate on the testing data would be near the desired .02. Likewise, in the subsampled dataset, a target Misclassification Rate of $q = .2$ was chosen.

\vspace{4mm} \noindent Using the full ozone data, we once again see some overfitting occurring on the training data. This can be seen as the training data has 0 misclassification error for all choices of $t$. Through the Thresholding Method, we once again select the threshold $t^*$ on the validation data curve. The closest misclassification error that still was below the cutoff on the validation data corresponded to yielded a misclassification error of .0196 and assigned 83\% of the points in the validation set. By using this $t^*$ to classify the testing data, we once again achieved an error below the threshold on the testing data. This corresponds to a satisfactory testing error of .016, while classifying 82\% of the points in the dataset.

\vspace{4mm} \noindent To demonstrate the flexibility of the Thresholding Method, the reduced ozone method shows that the Thresholding Method is independent of the method used to fit the data. The second method subsampled the data and only used the columns of the data most correlated with the labels instead of using the full dataset with all features. Regardless of this dimensionality reduction and subsampling, the Thresholding Method procedure functions identically.

\vspace{4mm} \noindent The reduced dataset exhibits strong overfitting for the training error and shows a poor fit compared to the full dataset. This also yields larger misclassification errors for all $t$. However, regardless of the additional dimensionality reduction step, we are still able to apply the Thresholding Method to target a maximum Misclassification Rate of .2. Using the validation data, this corresponds to $t^*$, yielding a Misclassification Rate of .197 on the validation data while assigning only 33\% of the points. When applied to the testing data, classifying with $t^* $ yields a satisfactory Misclassification Rate of .144 and classifies 38\% of the data.

\vspace{4mm} \noindent The Thresholding Method did not depend on the underlying method used to fit the mixture model. Therefore, the Thresholding Method can be compartmentalized and used after fitting to provide control over the error function of choice. This flexibility allows for an adaptable validation tool for use in many data analysis tasks.

\section{Conclusions}
\noindent The Thresholding Method provides a mechanism to exert error control in the testing data for a general black box classification scheme that outputs a set of labels and scores for each point. By parameterizing the problem through the sweep of a new tuning parameter $t$, we are able to identify which points are the most likely to be misclassified. By selecting a $t^*$ through the use a validation set such that our chosen error is below a target level, by classifying the new testing data with this $t^*$, the error rate in the testing set is controlled to be near that level.

\vspace{4mm} \noindent The strength of the Thresholding Method is exemplified through its flexibility. The method is able to be easily adapted to suit the underlying method being validated. The loss function of choice can be modified from Misclassification Rate to another validation metric such as Misclassification Loss. The only requirement on the loss function of choice is a weak monotonic assumption. Further, $t$ can be interpreted in many ways through means such as PA and AE. As long as the metric representing $t$ is monotonic, it will be satisfactory. This flexibility in the metrics of interest allows it to be easily compartmentalized and used in a variety of algorithms.

\vspace{4mm} \noindent The flexibility of the method is further seen in that it is agnostic to the underlying method used to fit the data. We demonstrated the method through the use of a Maximum Likelihood Classifier, but the framework can easily be modified to work in other classification schemes such as SVMs. The meaning of the threshold $t$ simply must be reparametrized to fit the method of interest.

\vspace{4mm} \noindent We further demonstrated the effectiveness of the method through simulation and real data results. The method provides control over given loss functions in the explored Gaussian Mixture Model scenario, though it can be adapted to work with many other data distributions. We further showed ways in which the use of the hold-out set provides improved performance over naively fitting the model with more data and no hold-out set.

\vspace{4mm} \noindent Future directions of interest include exploring the flexibility of the method further through application in other classification methods. In particular, we are interested in how different applications of the Thresholding Method using different loss functions may provide insight into establishing a theoretical backing for the method. With this theoretical foundation, we will study the magnitude of control provided by the method.

\end{document}